\documentclass[10pt,twocolumn,letterpaper]{article}

\usepackage{cvpr}
\usepackage{times}
\usepackage{epsfig}
\usepackage{graphicx}
\usepackage{amsmath}
\usepackage{amssymb}

\usepackage{amsthm}
\usepackage{xifthen}
\usepackage{xcolor}
\usepackage{colortbl}
\usepackage{wrapfig}
\usepackage{setspace} 
\usepackage{enumitem}
\usepackage{bbm}

\newcommand{\hl}[1]{\colorbox{yellow}{#1}}

\newcommand{\more}[1][]{%
\ifthenelse{\equal{#1}{}}
{\hl{\itshape more here\dots}\quad}
{\hl{\itshape more here on #1}\quad}
}
\newcommand{\note}[2][]{%
\textcolor{teal}{\sffamily\itshape\small%
\ifthenelse{\equal{#1}{}}
{} 
{#1: }
#2}
}

\DeclareMathOperator*{\argmin}{arg\,min}

\usepackage[sort]{natbib}
\bibpunct{[}{]}{,}{n}{}{}

\usepackage[pagebackref=true,breaklinks=true,letterpaper=true,colorlinks,bookmarks=false]{hyperref}

\cvprfinalcopy 

\def\cvprPaperID{372} 

\ifcvprfinal\fi
\begin{document}

\title{Actor-Action Semantic Segmentation with Grouping Process Models}

\author{Chenliang Xu and Jason J. Corso\\
Electrical Engineering and Computer Science\\
University of Michigan\\
{\tt\small \{cliangxu,jjcorso\}@umich.edu}
}

\maketitle

\begin{abstract}

Actor-action semantic segmentation made an important step toward advanced video 
understanding problems: what action is happening; who is performing the action; 
and where is the action in space-time. Current models for this problem are  
local, based on layered CRFs, and are unable to capture long-ranging interaction 
of video parts.  We propose a new model that combines these local labeling CRFs 
with a hierarchical supervoxel decomposition. The supervoxels provide cues for 
possible groupings of nodes, at various scales, in the CRFs to encourage 
adaptive, high-order groups for more effective labeling.  Our model is dynamic 
and continuously exchanges information during inference: the local CRFs 
influence what supervoxels in the hierarchy are active, and these active nodes 
influence the connectivity in the CRF; we hence call it a \emph{grouping process 
model}.  The experimental results on a recent large-scale video dataset show a 
large margin of 60\% relative improvement over the state of the art, which 
demonstrates the effectiveness of the dynamic, bidirectional flow
between labeling and grouping.


\end{abstract}


\section{Introduction}
\label{sec:intro}

Advances in modern high-level computer vision have helped usher in a new era of 
capable, perceptive physical platforms, such as automated vehicles.  As the 
performance of these systems improves, the expectations of their capabilities and 
tasks will also increase, commensurately, with platforms moving from the 
highways into our homes, for example.  
The need for these platforms to understand
not only \textbf{what} action is happening, but also \textbf{who} is doing the 
action and \textbf{where} the action is happening, will be increasingly critical 
to extracting semantics from video and, 
ultimately, to interacting with humans in our complex world. For example, a 
home kitchen robot must distinguish and locate \textit{adult-eating}, 
\textit{dog-eating} and \textit{baby-crying} in order to decide how to prepare 
and when to serve food. 


Despite the recent successes of aspects of this problem, such as action recognition~\cite{GoBlShTPAMI2007, ScLaCaICPR2004, SoZaSharXiv2012, KuJhGaICCV2011, LaIJCV2005, WaKlScIJCV2013, WaScICCV2013}, action segmentation~\cite{LuXuCoCVPR2015, JaVaJeCVPR2014}, video object segmentation~\cite{GiMuPaCVPR2015, LeKiGrICCV2011, ZhJaShECCV2014, PaFeICCV2013, LiKiHuICCV2013, OcMaBrTPAMI2014}, the collective problem had not been codified until~\cite{XuHsXiCVPR2015}, which posed a new actor-action semantic segmentation task on a large-scale YouTube video dataset called A2D. 
This dataset contains seven classes of 
actors including both articulated (e.g. \textit{baby}, \textit{cat} and \textit{dog}) 
and rigid (e.g. \textit{car} and \textit{ball}) ones, and eight classes of actions 
(e.g. \textit{flying}, \textit{walking} and \textit{running}). The task is to 
label each pixel in a video as a pair of actor and action labels or a null 
actor/action; one third of A2D videos contain multiple actors and actions.


\begin{figure*}[t!]
\centering
\includegraphics[width=0.85\linewidth]{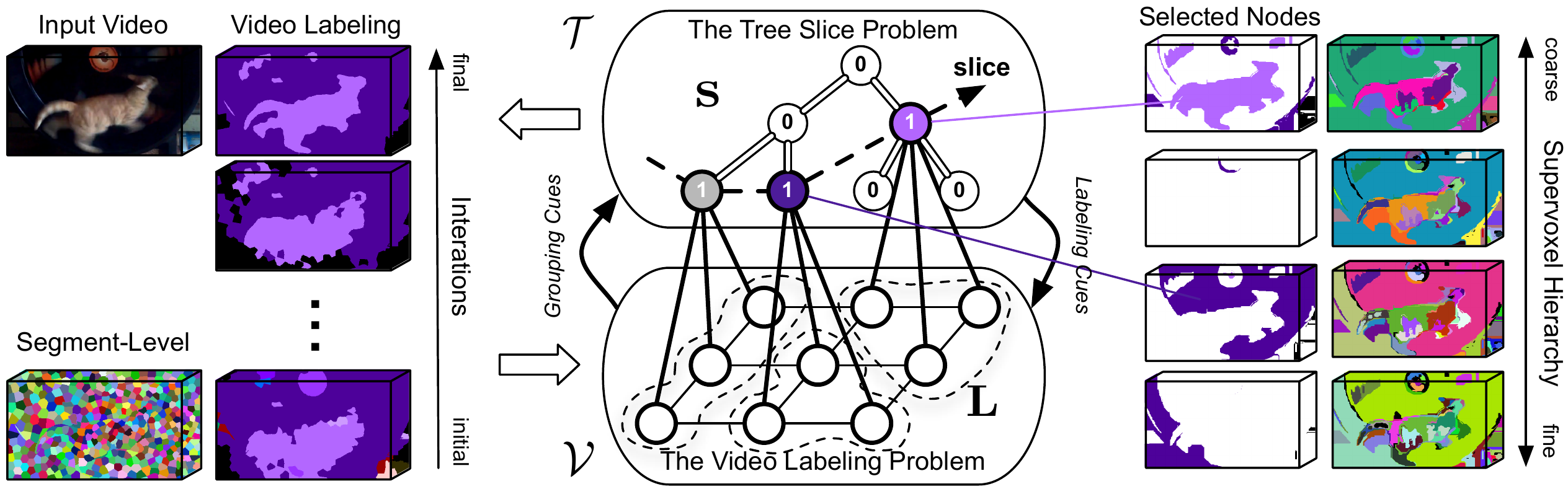}
\caption{Grouping Process Model with Bidirectional Inference. 
The local CRF at the segment-level starts with a coarse video labeling 
to influence what supervoxels in a hierarchy are active (grouping cue). 
The active supervoxels, in turn, influence the connectivity in the CRF,
thus refining the labels (labeling cue). This process is dynamic and 
continuous. The left side shows an example video with its segment-level segmentation
and its iteratively refined labels. The right side shows a supervoxel hierarchy and 
its active nodes.}
\label{fig:model}
\vspace{-4mm}
\end{figure*}

This task is challenging---their benchmarked leading method, the 
\textit{trilayer model}, only achieves a 26.46\% per-class pixel-level accuracy 
for  joint actor-action labeling. This model builds a large three-layer CRF on 
video supervoxels, where random variables of actor, actor-action, and action 
labels are defined on each layer, respectively, and connects layers with two 
sets of potential functions that capture conditional probabilities (e.g.  
conditional distribution of action given a specific actor class). Although their 
model accounts for the interplay of actors and actions, the interactions of the two 
sets of labels are restricted to the local CRF neighborhoods, which, based on 
the low absolute performance they achieve, is
insufficient to solve this unique actor-action problem for three reasons. 

First, we believe the pixel-level model must be married to a secondary process 
that captures instance-level or video-level global information, such as action 
recognition, in order to properly model the actions. Lessons learned from 
images strongly supports this argument---the performance of semantic image 
segmentation on the MSRC dataset seems to hit a plateau \cite{ShWiRoECCV2006} until 
information from secondary processes, such as context \cite{LaRuKoTPAMI2014, 
MoChLiCVPR2014}, object detectors \cite{LaStAlECCV2010} and a holistic scene 
model \cite{YaFiUrCVPR2012}, are added. However, to the best of our knowledge, 
there is no method in video semantic segmentation that directly leverages the 
recent success in action recognition. 



Second, these two sets of labels, actors and actions, exist at different levels 
of granularities. For example, suppose we want to label \textit{adult-clapping} in a 
video. The actor, \textit{adult}, can probably be recognized and labeled by looking 
only at the lower body, e.g. legs. However, in order to recognize and label the 
\textit{clapping} action, we have to either localize the acting parts of the 
human body or simply look at the whole actor body for recognizing the action.

Third, actors and actions have different orientations along space and time 
dimensions in a  video. Actors are more space-oriented---they can be fairly well 
labeled using only still images, as in semantic image 
segmentations~\cite{ShWiRoIJCV2009, YaFiUrCVPR2012}, whereas actions are more 
space-time-oriented. Although one can possibly identify actions by still images 
alone~\cite{YaJiKhICCV2011}, there are strong 
distinctions between different actions along the time dimension.  For example, 
\textit{running} is faster and has more repeated motion patterns than 
\textit{walking} for a given duration; and \textit{walking} performed by a 
\textit{baby} is very different compared to an \textit{adult}, although they may 
easily confuse a spatially trained object detector, such as DPM~\cite{FeGiMcTPAMI2010}, 
without more complex spatiotemporal modeling.


Our method overcomes the above limitations in two ways: (1) we propose 
a novel grouping process model (GPM) that adaptively groups segments together 
during inference, and (2) we incoporate video-level recognition into 
segment-level labeling thru multi-label labeling costs and the grouping process model. 
The GPM is a dynamic and continuous process of information exchange 
of the labeling CRFs and a supervoxel hierarchy. The supervoxel hierarchy 
provides a rich multi-resolution decomposition of the video content, where 
object parts, deformations, identities and actions are retained in space-time 
supervoxels across multiple levels in the hierarchy \cite{XuWhCoICCV2013, 
JaVaJeCVPR2014, OnReVeECCV2014}. Rather than using object and action proposals 
as separate processes, we directly localize the actor and action nodes in a 
supervoxel hierarchy by the labeling CRFs. During inference, the labeling CRFs 
influence what supervoxels in a hierarchy are \textit{active}, and these active 
supervoxels, in turn, influence the connectivity in the CRF, thus refining the 
labels. 

This bidirectional inference for GPM is dynamic and iterative as shown in 
Fig.~\ref{fig:model} and can be efficiently solved by graph cuts and binary 
linear programming. We show that the GPM can be effectively combined with 
video-level recognition signals to efficiently influence the actor-action 
labelings in video segmentation. Throughout the entire inference process, the 
actor and action labels exchange information at various levels in the 
supervoxel hierarchy, such that the multi-resolution and space-time orientations 
of the two sets of labels are explicitly explored in our model.

We conduct thorough experiments on the large-scale actor-action video dataset 
(A2D) \cite{XuHsXiCVPR2015}. We compare the 
proposed method to the previous benchmarked leading method, the trilayer model, 
as well as two leading semantic segmentation methods~\cite{LaRuKoTPAMI2014, 
KrKoNIPS2011} that we have extended to the actor-action problem. The 
experimental results show that our proposed method, which is driven by the 
grouping process model, outperforms the second best method by a large margin of 
17\% per-class accuracy  (60\% relative improvement) and over 10\% global pixel 
accuracy, which demonstrates the effectiveness of our modeling.


\vspace{-2mm}
\section{Related Work}
\label{sec:related}

Our paper is closely related to Xu et al. \cite{XuHsXiCVPR2015}, where the 
actor-action semantic segmentation problem is first proposed. Their paper 
demonstrates that inference jointly over actors and actions outperforms 
inference independently over them.  They propose a trilayer model that achieves 
the state-of-the-art performance on the actor-action semantic segmentation 
problem.  However, their model only captures the interactions of actors and 
actions in a local CRF pairwise neighborhood, whereas our method considers the 
interplays at various levels of granularities in space and time introduced by a 
supervoxel hierarchy. 

Supervoxels have demonstrated potential to capture object boundaries, follow 
object parts over time \cite{XuCoCVPR2012}, and localize objects and actions  \cite{JaVaJeCVPR2014,OnReVeECCV2014}.
Supervoxels are used as higher-order potentials for human action 
segmentation~\cite{LuXuCoCVPR2015} and video object segmentation~\cite{JaGrECCV2014}. 
Different from the above works, we use a supervoxel hierarchy to connect 
bottom-up pixel labeling and top-down recognition, where supervoxels contain 
clear actor-action semantic meaning. We also use the tree slice concept for 
selecting supervoxels in a hierarchy as in \cite{XuWhCoICCV2013}, but the 
difference is that our model selects the tree slices in an iterative fashion, 
where the tree slice also modifies the pixel-level groupings.


Our work also differs from the emerging works in action localization, action 
detection, and video object segmentation for two reasons. First, our 
segmentation contains clear semantic meanings of actor and action labels, whereas 
most existing works in action localization and detection do not~\cite{JaVaJeCVPR2014, MaZhIkICCV2013, TrYuNIPS2012, LaWaMoICCV2011}. Second, we 
consider multiple actors performing actions in a video and explicitly model the 
types of actors, whereas existing works assume one human actor~\cite{MaSiScCVPR2015, YuYuCVPR2015, TiSuShCVPR2013, YuLiWuCVPR2009} 
or do not model the types of 
actors at all~\cite{GiMuPaCVPR2015, LeKiGrICCV2011, ZhChLiCVPR2015, 
ZhJaShECCV2014}.  Although there have been some works on action detection~\cite{TiSuShCVPR2013}, this remains an open challenge.

We relate our work to AHRF \cite{LaRuKoTPAMI2014} and FCRF~\cite{KrKoNIPS2011} 
in Section \ref{sec:infer} after presenting the new model.

\section{Grouping Process Model}
\label{sec:gpm}

Grouping Process Model (GPM) is a dynamic and continuous process of information exchange during inference: the local CRF influences what supervoxels in a hierarchy are active, and these active supervoxels, in turn, influence the connectivity in the CRF. Here, we give its general form, and Fig.~\ref{fig:model} shows an overview. We define the detailed potentials adapted to the actor-action problem in Sec.~\ref{sec:model}.

\noindent \textbf{Segment-level.} Without loss of generality, we define 
$\mathcal{V} = \{q_1, q_2, \dots, q_N\}$ as a video with $N$ voxels or a video 
segmentation with $N$ segments. A graph $\mathcal{G} = (\mathcal{V}, 
\mathcal{E})$ is defined over the entire video, where the neighborhood structure of 
the graph $\mathcal{E}(\cdot)$ is induced by the connectivities in the voxel 
lattice $\Lambda^3$ or segmentation graph over space-time in a video. We 
define a set of random variables $\mathbf{L} = \{l_1, l_2, \dots, l_N\}$ where 
the subscript corresponds to a certain node in $\mathcal{V}$ and each $l_i \in 
\mathcal{L}$ takes some label from a label set.  The GPM is inherently a 
labeling CRF, but it leverages a supervoxel hierarchy to 
dynamically adjust its non-local grouping structure.

\noindent \textbf{Supervoxel Hierarchy.} Given a coarse-to-fine supervoxel 
hierarchy generated by a hierarchical video segmentation method, such as GBH 
\cite{GrKwHaCVPR2010}, we extract a supervoxel tree\footnote{We add one 
virtual node as root to make it a tree if the segmentation at the coarsest level 
contains more than one segment.}, denoted as $\mathcal{T}$ with $\mathcal{S}$ 
total nodes in the tree, by ensuring that each supervoxel at a finer level 
segmentation has one and only one parent at its coarser level 
(Sec.~\ref{sec:exp} details the tree extraction process in the general case).
We define a set of random variables $\mathbf{s} = \{s_1, s_2, \dots, 
s_\mathcal{S}\}$ on supervoxel nodes in the entire tree 
$\mathcal{T}$, where $s_t \in \{0,1\}$ takes a binary label to indicate 
whether the $t$th supervoxel node is active or not. Being a segmentation 
hierarchy, each supervoxel connects to a set of segment nodes by their
overlapping in voxel lattice $\Lambda^3$, thus we have
each $s_t$ connecting to a set of random variables at the segment level, 
denoted as $\mathbf{L}_t$.


Supervoxel hierarchies, such as \cite{GrKwHaCVPR2010, CoShDuTMI2008}, 
are built by iteratively recomputing and merging finer supervoxels into coarser 
ones based on appearance and motions, where the body parts of 
an actor and its local action are contained at the finer levels in the hierarchy and 
the identity of the actor and its long-ranging action are contained at the coarser levels. 
But going too coarse will cause oversegmentation with the background and going too fine will lose the meaningful actions. Therefore it is challenging to locate the  
supervoxels in a hierarchy that best describe the actor and its action. 
Instead, our GPM uses the evidence from a second 
source---the segment-level CRF, to locate the supervoxels 
supported by the labeling $\mathbf{L}$. Once the supervoxels $\mathbf{s}$ are 
selected, they provide strong labeling cues to the segment-level 
labeling---these segment-level nodes are from a same actor or a 
same action, thus they can be fully connected to refine the labeling. 

The objective of GPM is to find the best labeling of $\mathbf{L}^*$ and 
$\mathbf{s}^*$ to minimize the following energy:
\begin{align}
(\mathbf{L}^*, \mathbf{s}^*) &= \argmin_{\mathbf{L}, \mathbf{s}} 
E(\mathbf{L},\mathbf{s}|\mathcal{V},\mathcal{T}) 
\nonumber\\
E(\mathbf{L},\mathbf{s}|\mathcal{V},\mathcal{T}) 
&= E^{v}(\mathbf{L}|\mathcal{V}) + E^h(\mathbf{s}|\mathcal{T})
\label{eq:energy_gpm}  \\
&+ \sum_{t\in\mathcal{S}} (E^{h}(\mathbf{L}_t|s_t) + 
E^{h}(s_t|\mathbf{L}_t))
\nonumber \enspace,
\end{align}
where $E^{v}(\mathbf{L}|\mathcal{V})$ and $E^{h}(\mathbf{s}|\mathcal{T})$ 
encode the potentials at the segment-level and the supervoxel hierarchy, 
respectively; $E^h(\mathbf{L}_t|s_t)$ and $E^{h}(s_t|\mathbf{L}_t)$ 
are conditional potential functions defined as directional edges in 
Fig.~\ref{fig:model}.  To keep the discussion general, we do not define the 
specific form of $E^{v}(\mathbf{L}|\mathcal{V})$ here, it can be any 
segment-level CRF, such as~\cite{LaRuKoTPAMI2014, KrKoNIPS2011, ShWiRoECCV2006}. We define the other terms next. 

\subsection{Labeling Cues from Supervoxel Hierarchy}

Given an active node $s_t$ in the supervoxel hierarchy, we use it as a cue to refine the segment-level labelings and we define the energy of this process as:
\begin{align}
E^{h} (\mathbf{L}_t|s_t)
= \left \{\begin{array}{ll}
\sum_{i\in\mathbf{L}_t} \sum_{j \neq i}
\psi_{ij}^{h}(l_i, l_j) & \mbox{ if $s_t=1$}\\
0 & \text{otherwise.}\end{array}
\right.
\label{eq:energy_gpm_labeling}
\end{align}
Here, $\psi_{ij}^{h}(\cdot)$ has the form:
\begin{align}
\psi_{ij}^{h}(l_i, l_j) 
= \left \{\begin{array}{ll}
\theta_t & \mbox{ if $l_i \neq l_j$}\\
0 & \text{otherwise,}\end{array}
\right.
\end{align}
where $\theta_t$ is a constant parameter to be learned. $\psi_{ij}^{h}(l_i, l_j)$ penalizes any two nodes in the field $\mathbf{L}_t$ that contain different labels. Eq.~\ref{eq:energy_gpm_labeling} will change the graph structure in  $\mathbf{L}_t$ by fully connecting the nodes inside, and has clear semantic meaning----this set of nodes in $\mathbf{L_t}$ at the segment-level are linked to the same supervoxel node $s_t$ and hence they are from the same object, taking evidences from the appearance and motion features used in a typical supervoxel segmentation method.

\subsection{Grouping Cues from Segment Labeling}


If the selected supervoxels are too fine, they are subject to 
losing object identity and long-ranging actions; if they are too coarse, 
they are subject to oversegmenting with the background. 
Therefore, we set the selected supervoxels to best reflect the 
segment-level labelings while also respecting a selection prior.
Given some video labeling at the segment-level, we select the nodes in the 
supervoxel hierarchy that best correspond to this current labeling: 
\begin{align}
E^{h} (s_t|\mathbf{L}_t) = (\mathcal{H}(s_t) |s_t| + \theta_h) s_t
\enspace,
\label{eq:energy_gpm_grouping}
\end{align}
where $|\cdot|$ denotes the size of a supervoxel in terms of video voxels and $\theta_h$ is a 
parameter to be learned that encodes a prior of the node selection in the 
hierarchy. $\mathcal{H}(\cdot)$ is defined as the entropy of the labeling 
field connected to $s_t$:
\begin{align}
\mathcal{H}(s_t) = -\sum_{\gamma\in\mathcal{L}} 
P(\gamma; \mathbf{L}_t) \log P(\gamma; \mathbf{L}_t)
\enspace,
\label{eq:entropy}
\end{align}
where $P (\gamma; \mathbf{L}_t) = \frac{\sum_{i\in\mathbf{L}_t} 
\delta(l_i=\gamma)}{|\mathbf{L}_t|}$ and $\delta(\cdot)$ is an indicator 
function. Intuitively, the first term in Eq.~\ref{eq:energy_gpm_grouping} 
pushes down the selection of nodes in the hierarchy such that they only include 
the labeling field that has same labels, and the second term pulls up the node 
selection, giving penalties for going down the hierarchy. 

\subsection{Valid Active Nodes By The Tree Slice}

The active nodes in $\mathbf{s}$ define what groups of segments the GPM will 
enforce during labeling; hence the name grouping process model.  However, not 
all active node sets $\mathbf{s}$ are permissible: since we seek a single 
labeling over the video, we enforce that each node in $\mathcal{V}$ (each 
segment) is associated with one and only one active group in 
$\mathbf{s}$.  This notion was introduced in \cite{XuWhCoICCV2013} by way of a 
\textit{tree slice}: on every root-to-leaf path in the tree $\mathcal{T}$ one 
and only one node in $\mathbf{s}$ is active.  

We follow \cite{XuWhCoICCV2013} to define a matrix $\mathcal{P}$ that 
encodes all root-to-leaf paths in $\mathcal{T}$. $\mathbf{P}_p$ is one row in $\mathcal{P}$, and 
it encodes the path from the root to $p$th leaf with $1$s for nodes on the path and 
$0$s otherwise.
We define the energy to regulate $\mathbf{s}$ as:
\begin{align}
E^h(\mathbf{s}|\mathcal{T}) = \sum_{p=1}^{P} \delta(\mathbf{P}_p \cdot 
\mathbf{s} \neq 1) \theta_\tau
\enspace,
\label{eq:energy_gpm_slice}
\end{align}
where $P$ is the total number of leaves (also the number of such root-to-leaf paths), 
$\cdot$ denotes dot product, and $\theta_\tau$ is a large constant to 
penalize an invalid tree slice. The tree slice selects supervoxel nodes to form a 
new video representation that has a one-to-one mapping to the video 3D 
lattices $\Lambda^3$.


\section{Bidirectional Inference for GPM}
\label{sec:infer}


In this section, we show that we can use an iterative bidirectional inference schema to efficiently solve the objective function defined in Eq.~\ref{eq:energy_gpm}---given the segment-level labeling, we find the best supervoxels in the hierarchy; and given the selected supervoxels in the hierarchy, we regroup the segment-level labeling.

\noindent \textbf{The Video Labeling Problem.} Given a tree slice $\mathbf{s}$, we would like to find the best $\mathbf{L}^*$. Formally, we have:
\begin{align}
&\mathbf{L}^* = \argmin_\mathbf{L} E(\mathbf{L}|\mathbf{s}, \mathcal{V}) 
\label{eq:infer_bottom} \\
&\quad = \argmin_\mathbf{L} E^{v}(\mathbf{L}|\mathcal{V}) 
+ \sum_{t\in\mathcal{S}} E^{h}(\mathbf{L}_t|s_t)
\nonumber
\enspace.
\end{align}
This equation can have a standard CRF form depending on how $E^{v}(\mathbf{L}|\mathcal{V})$ is defined. The higher-order energy we defined in $E^h(\mathbf{L}_t|s_t)$ can be decomposed to a locally fully connected CRF, and its range is constrainted by $s_t$ such that the inference is inexpensive even without Gaussian kernels~\cite{KrKoNIPS2011}. 

\noindent \textbf{The Tree Slice Problem.} Given the current labeling 
$\mathbf{L}$, we would like to find the best $\mathbf{s}^*$. Formally, we 
have:
\begin{align}
&\mathbf{s}^* = \argmin_{\mathbf{s}} E(\mathbf{s}|\mathbf{L},\mathcal{V}) 
\label{eq:infer_top} \\
&\quad = \argmin_{\mathbf{s}} E^{h}(\mathbf{s}|\mathcal{T}) + \sum_{t\in\mathcal{S}} 
E^{h}(s_t|\mathbf{L}_t)
\nonumber
\enspace.
\end{align}
We use binary linear programming to optimize Eq.~\ref{eq:infer_top}, and thus we 
rewrite the problem to have the following form: 
%
\begin{align}
\min \sum_{t\in\mathcal{S}} \alpha_t s_t \quad\text{s.t.}\;
\mathcal{P} \mathbf{s} = \mathbf{1}_P \;\;\text{and}\;\;
\mathbf{s} = \{0,1\}^{\mathcal{S}}\enspace,
\label{eq:blp_objective} 
\end{align}
where $\alpha_t = \mathcal{H}(s_t) |s_t| + \theta_h$. Note that this optimization is 
substantially simpler than that proposed by the original tree slice paper 
\cite{XuWhCoICCV2013}, which incorporated quadratic terms in a binary quadratic 
program.  We use a standard solver (IBM CPLEX) to solve the binary linear 
programming problem.  

\noindent \textbf{Iterative Inference.} The above two conditional inferences are 
iteratively carried out, as depicted in Fig.~\ref{fig:model}. To be specific, we 
initialize a coarse labeling $\mathbf{L}$ by solving Eq.~\ref{eq:infer_bottom} 
without the second term, then we solve 
Eq.~\ref{eq:infer_top} and \ref{eq:infer_bottom} in an iterative fashion.  
Each round of the tree slice problem enacts an updated set of grouped segments, 
which are then encouraged to be assigned the same label during the subsequent 
labeling process.  Although we do not include a proof of convergence in this 
paper, we notice that the solution converges after a few rounds.


\noindent \textbf{Relation to AHRF.} The associative hierarchical random field 
(AHRF) \cite{LaRuKoTPAMI2014} implicitly pushes up the inference nodes towards 
higher-levels in the segmentation tree $\mathcal{T}$, whereas our model (GPMs) 
explicitly models the best set of active nodes in the segmentation tree 
$\mathcal{T}$ by the means of a tree slice. AHRF defines a full random field on 
the hierachy; our model leverages the hierarchy to adaptively group at the 
pixel level.  Our model is hence more scalable to videos. 
GPMs assume that the best 
representations of the video content exist in a tree slice rather than 
enforcing the agreement across different levels as in AHRF. For example, a 
video of \textit{long jumping} often contains \textit{running} in 
the beginning. The running action exists and has a strong classifier signal at 
a fine-level in a supervoxel hierarchy, but it quickly diminishes when one 
goes to a higher level in the hierarchy where supervoxels capture longer range 
time dimension in the video and would then favor the jumping action.

\noindent \textbf{Relation to FCRF.} The fully-connected CRF (FCRF) 
in~\cite{KrKoNIPS2011} imposes a Gaussian mixture kernel to regularize the 
pairwise interactions of nodes. Although our model fully connects the nodes in 
each $\mathbf{L}_t$ for a given iteration of inference, we explicitly take the 
evidence from the supervoxel groupings rather than a Gaussian kernel. The 
energy in Eq.~\ref{eq:energy_gpm_grouping} restricts the selected supervoxels to 
avoid overmerging. Although a more complex process, in practice, our inference is 
efficient. It takes on the order of seconds for a typical video with a few 
thousand label nodes and a few hundred supervoxel nodes.

\section{The Actor-Action Problem Modeling}
\label{sec:model}

Typical semantic segmentation methods \cite{XuHsXiCVPR2015, TiLaIJCV2012} train 
classifiers at the segment-level. In our case, these segment-level classifiers 
capture the local appearance and motion features of the actors' body parts; 
they have some ability to locate the actor-action in a video, but these
predictions are noisy since they do not capture the actor-whole or leverage any 
context information in the video.  Video-level classifiers, as a secondary 
process, capture the global information of actors performing actions and have 
good prediction performance at the video-level.  However, they are not able to 
localize where the action is happening. These two streams of 
information captured at the segment-level and at the video-level are 
complementary to each other.   In this section, we implement these two streams 
together in a single model, leveraging the grouping process model as a means of 
marrying the video-level signal to the segment level problem.

Let us first define notation, extending that from Sec.~\ref{sec:gpm} where 
possible.
We use $\mathcal{X}$ to denote the set of actor labels (e.g. \textit{adult}, 
\textit{baby} and \textit{dog}) and $\mathcal{Y}$ to denote the set of action 
labels (e.g. \textit{eating}, \textit{walking} and \textit{running}). The 
segment-level random fields $\mathbf{L}$ now take two labels---for the $i$th 
segment, $l_i^{\mathcal{X}} \in \mathcal{X}$ is a label from the actor set and 
$l_i^{\mathcal{Y}} \in \mathcal{Y}$ is a label from the action set. 
%
%
We define $\mathcal{Z}=\mathcal{X}\times\mathcal{Y}$ as the joint product space 
of the actor and action labels. We define a set of binary random variables 
$\mathbf{v} = \{v_1,v_2,\dots,v_{|\mathcal{Z}|}\}$ on the video-level, where 
$v_z=1$ denotes the $z$th actor-action label is active at the video-level.
They represent the video-level multi-label labeling problem. 
Again, we have the set of binary random variables $\mathbf{s}$ defined on the 
supervoxel hierarchy as in Sec.~\ref{sec:gpm}.

Therefore, we have the total energy function of the actor-action semantic segmentation defined as:
\begin{align}
&(\mathbf{L}^*, \mathbf{s}^*, \mathbf{v}^*) = 
\argmin_{\mathbf{L},\mathbf{s},\mathbf{v}}
E(\mathbf{L},\mathbf{s},\mathbf{v}|\mathcal{V},\mathcal{T}) 
\nonumber \\
&E(\mathbf{L},\mathbf{s},\mathbf{v}|\mathcal{V},\mathcal{T}) 
= E^{v}(\mathbf{L}|\mathcal{V}) 
+ \sum_{z\in\mathcal{Z}} E^\mathcal{V}(v_z|\mathcal{V})
+ E^\mathcal{V}(\mathbf{L},\mathbf{v})
\nonumber \\
&\quad + E^h(\mathbf{s}|\mathcal{T}) 
+ \sum_{t\in\mathcal{S}} 
(E^{h} (\mathbf{L}_t,\mathbf{v}|s_t) + E^{h} (s_t|\mathbf{L}_t))
\label{eq:energy_total}
\enspace,
\end{align}
where the term $E^{h} (\mathbf{L}_t,\mathbf{v}|s_t)$ now 
models the joint potentials of the segment-level labeling field $\mathbf{L}_t$ 
and the video-level labeling $\mathbf{v}$, which is slightly different from its 
form in Eq.~\ref{eq:energy_gpm_labeling}. We have two new terms, 
$E^\mathcal{V}(v_z|\mathcal{V})$ and $E^\mathcal{V}(\mathbf{L},\mathbf{v})$, 
from the video-level, where $v_z$ is the $z$th coordinate in $\mathbf{v}$. 
We explain these new terms next.

\subsection{Segment-Level CRF $E^v$}
\label{sec:model_segment}


At the segment-level, we use the same bilayer actor-action CRF model 
from \cite{XuHsXiCVPR2015} to capture the local pairwise interactions of the two sets of labels:
%
\begin{align}
&E^{v}(\mathbf{L}|\mathcal{V}) 
= \sum_{i\in\mathcal{V}} \psi_i^v (l_i^\mathcal{X}) 
+ \sum_{i\in\mathcal{V}}\sum_{j\in\mathcal{E}(i)} \psi_{ij}^v (l_i^\mathcal{X},l_j^\mathcal{X}) 
\label{eq:energy_seg} \\
&+ \sum_{i\in\mathcal{V}} \phi_i^v (l_i^\mathcal{Y}) 
+ \sum_{i\in\mathcal{V}}\sum_{j\in\mathcal{E}(i)} \phi_{ij}^v (l_i^\mathcal{Y},l_j^\mathcal{Y}) 
+ \sum_{i\in\mathcal{V}} \varphi_i^v (l_i^\mathcal{X}, l_i^\mathcal{Y})
\nonumber \enspace,
\end{align}
where $\psi_i^v$ and $\phi_i^v$ encode separate potentials for a random variable 
$l_i$ to take the actor and action labels, respectively. $\varphi_i^v$ is a 
potential to measure the compatibility of the actor-action tuples on segment 
$i$, and $\psi_{ij}^v$ and $\phi_{ij}^v$ capture the pairwise interactions 
between segments, which have the form of a contrast sensitive Potts 
model \cite{BoJoICCV2001, ShWiRoECCV2006}. We use the publicly available code from \cite{XuHsXiCVPR2015} to capture the local pairwise interactions of the two sets of labels.

%

\subsection{Video-Level Potentials $E^\mathcal{V}$}
\label{sec:model_video}

Rather than a uniform penalty over all labels \cite{DeOsIsIJCV2012}, 
we use the video-level recognition signals as global multi-label labeling costs to impact 
the segment-level labeling. We define the unary energy at the video-level as:
\begin{align}
E^\mathcal{V}(v_z|\mathcal{V}) = -(\xi^\mathcal{V} (z)-\theta_T) \theta_B v_z
\label{eq:video_multi_unary}
\enspace,
\end{align}
where $\xi^\mathcal{V} (\cdot)$ is the video-level classification response for 
a particular actor-action label, and Sec.~\ref{sec:exp} describes its training process. 
Here, $\theta_T$ is a parameter to control response threshold, and $\theta_B$ is 
a large constant parameter. In other words, to minimize Eq.~\ref{eq:video_multi_unary}, 
the label $v_z=1$ only when the classifier response 
$\xi^\mathcal{V} (z)>\theta_T$.

We define the interactions between the video-level and the segment-level:
\begin{align}
E^{\mathcal{V}}(\mathbf{L},\mathbf{v}) = 
\sum_{x\in\mathcal{X}} \delta_{x} (\mathbf{L}) 
h_{x} (\mathbf{v}) \theta_\mathcal{V}
+ \sum_{y\in\mathcal{Y}} \delta_{y} (\mathbf{L}) 
h_{y} (\mathbf{v}) \theta_\mathcal{V}
\label{eq:energy_video_seg}
\enspace,
\end{align}
where $\delta_x(\cdot)$ is an indicator function to determine whether the current labeling $\mathbf{L}$ at the segment-level contains a particular label $x \in \mathcal{X}$ or not:
\begin{align}
\delta_x(\mathbf{L}) = \left \{\begin{array}{ll}
1 & \mbox{if $\exists i : l_i^\mathcal{X} = x$}\\
0 & \text{otherwise}.\end{array}
\right.
\end{align}
Here, $h_{x} (\cdot)$ is another indicator function to determine whether 
a particular label $x$ is supported at the video-level or not:
\begin{align}
h_{x} (\mathbf{v}) = \left \{\begin{array}{ll}
0 & \mbox{if $\exists z : v_z=1 \wedge g(z)=x$ }\\
1 & \text{otherwise},\end{array}
\right.
\end{align}
where $g(\cdot)$ maps a label in the joint actor-action space to the actor space.
$\theta_{\mathcal{V}}$ is a constant cost for any label that exists in 
$\mathbf{L}$ but not supported at the video-level.
We define $\delta_y(\cdot)$ and $h_{y} (\cdot)$ similarly. 
To make the cost meaningful, we set $\theta_B>2\theta_\mathcal{V}$.
In practice, we observe that these labeling costs 
from video-level potentials help the segment-level labeling to achieve a more 
parsimonious-in-labels result that enforces more global information than using 
local segments along (see results in Table \ref{tab:exp_dataset}).

\subsection{The GPM Potentials $E^h$}
\label{sec:model_tree}

\begin{figure}[t]
\centering
\includegraphics[width=0.99\linewidth]{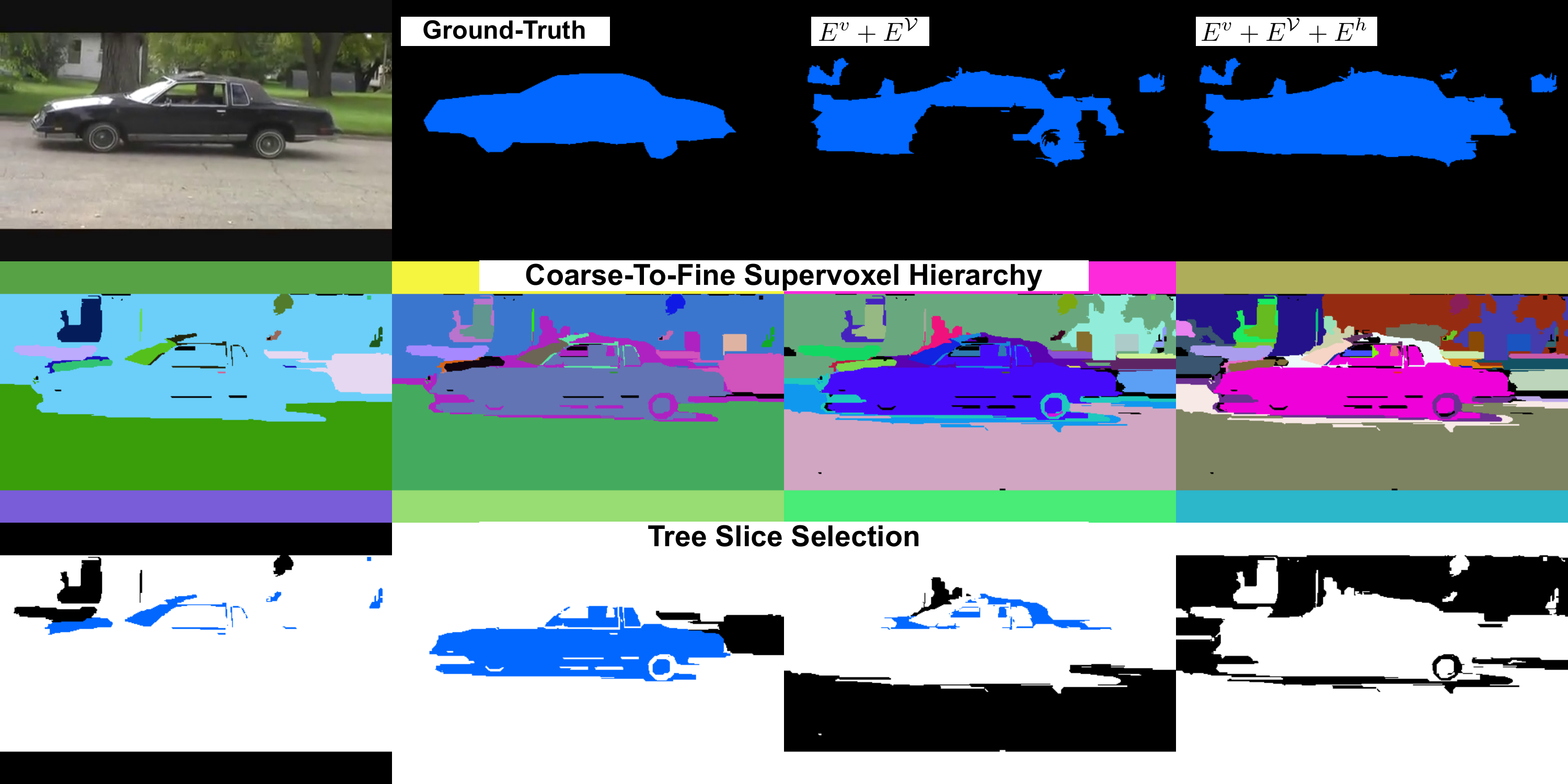}
\caption{Actor-action video labeling is refined by GPM. First row shows 
a test video \textit{car-jumping} with its labelings. The second row shows the 
supervoxel hierarchy and the third row shows the active nodes with their 
dominant labels.}
\label{fig:visual_gbh}
\end{figure}


The energy terms $E^h(\mathbf{s}|\mathcal{T})$  and $E^{h} (s_t|\mathbf{L}_t)$ 
involved in the tree slice problem are defined the same as in Sec.~\ref{sec:gpm}. 
Now, we define the new labeling term:
\begin{align}
&E^{h} (\mathbf{L}_t,\mathbf{v}|s_t) = 
\label{eq:energy_tree_seg} \\
&\quad \left \{\begin{array}{ll}
\sum_{i\in\mathbf{L}_t} \sum_{j \neq i}
\psi_{ij}^{h}(l_i^\mathcal{X}, l_j^\mathcal{X},\mathbf{v}) \\
\quad + \sum_{i\in\mathbf{L}_t} \sum_{j \neq i}
\phi_{ij}^{h}(l_i^\mathcal{Y}, l_j^\mathcal{Y},\mathbf{v}) & \mbox{if $s_t=1$}\\
0 & \text{otherwise.}\end{array}
\right. \nonumber
\end{align}
Here, $\psi_{ij}^{h}(\cdot)$ has the form:
\begin{align}
&\psi_{ij}^{h}(l_i^\mathcal{X}, l_j^\mathcal{X},\mathbf{v}) = \\
&\quad \left \{\begin{array}{ll}
\theta_t & \mbox{ if $l_i^\mathcal{X} \neq l_j^\mathcal{X}$, 
$\exists z : v_z=1 \wedge g(z)=f (s_t)$}\\
0 & \text{otherwise,}\end{array}
\right.
\nonumber
\end{align}
where $f(\cdot)$ denotes the dominant actor label in the segment-level labeling field $\mathbf{L}_t$ that connected to $s_t$, and we define $\psi_{ij}^{h}(l_i^\mathcal{Y},l_j^\mathcal{Y},\mathbf{v})$ similarly. This new term selectively refines the segmentation where the majority of the segment-level labelings agree with the video-level multi-label labeling. 

We show in Fig.~\ref{fig:visual_gbh} how this GPM process helps to refine the 
actor's shape (the car) in the segmentation labeling. The initial labelings from 
$E^v+E^\mathcal{V}$ propose a rough region of interest, but they do not capture 
the accurate boundaries and shape. After two iterations of inferences, 
the tree slice selects the best set of nodes in the GBH hierarchy that 
represents the actor, and they regroup the segment-level labelings such that the 
labelings can better capture the actor shape. Notice that the car body in the 
third column merges with the background, but our full model (fourth column) overcomes the limitation by selecting different parts from the hierarchy to yield the final 
grouping segmentation.

\subsection{Inference}

\begin{figure}[t]
\centering
\includegraphics[width=0.65\linewidth]{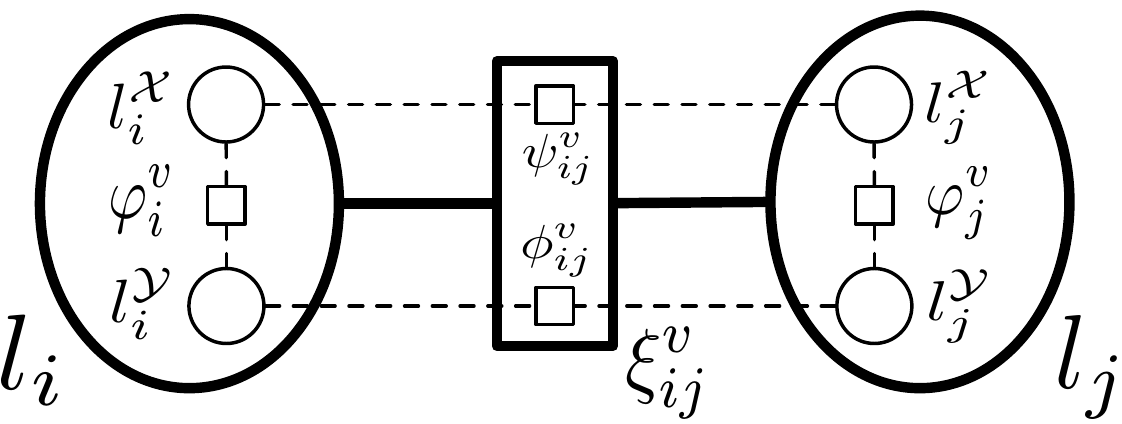}
\caption{Visualization of two nodes of the bilayer model in our efficient inference.}
\label{fig:infer_gco}
\vspace{-2mm}
\end{figure}

The inference of the actor-action problem defined in Eq.~\ref{eq:energy_total} 
follows the bidirectional inference described in Sec.~\ref{sec:infer}. The tree 
slice problem can be efficiently solved by binary linear programming. The video 
labeling problem could be solved using loopy belief propagation.  However, given 
the fact that the CRFs are defined over two sets of labels, the actors and 
actions, this inference problem would be very expensive.  Here, we derive a way 
to solve it efficiently using graph cuts inference with label costs 
\cite{BoKoTPAMI2004,BoVeZaTPAMI2001, DeOsIsIJCV2012}. 
We show this conceptually in Fig.~\ref{fig:infer_gco} and rewrite Eq.~\ref{eq:energy_seg} as:
\begin{align}
E^{v}(\mathbf{L}|\mathcal{V}) = \sum_{i\in\mathcal{V}} \xi_i^v (l_i) + 
\sum_{i\in\mathcal{V}} \sum_{j\in\mathcal{E}(i)} \xi_{ij}^v (l_i, l_j)
\label{eq:gc_seg_unary}
\enspace,
\end{align}
where we define the new unary as: 
\begin{align}
\xi_i^v (l_i) = \psi_i^v(l_i^\mathcal{X}) + \phi_i^v(l_i^\mathcal{Y}) + 
\varphi_i^v(l_i^\mathcal{X},l_j^\mathcal{Y})
\enspace,
\end{align}
and the pairwise interactions as:
\begin{align}
&\xi_{ij}^v (l_i,l_j) = \label{eq:gc_seg_binary} \\
&\left \{\begin{array}{ll}
\psi_{ij}^v (l_i^\mathcal{X}, l_j^\mathcal{X}) & \mbox{if $l_i^\mathcal{X} \neq 
l_j^\mathcal{X} \wedge l_i^\mathcal{Y} = l_j^\mathcal{Y}$} \\
\phi_{ij}^v (l_i^\mathcal{Y}, l_j^\mathcal{Y}) & \mbox{if $l_i^\mathcal{X} = 
l_j^\mathcal{X} \wedge l_i^\mathcal{Y} \neq l_j^\mathcal{Y}$} \\
\psi_{ij}^v (l_i^\mathcal{X}, l_j^\mathcal{X}) \phi_{ij}^v (l_i^\mathcal{Y}, 
l_j^\mathcal{Y}) & \mbox{if $l_i^\mathcal{X} \neq l_j^\mathcal{X} \wedge l_i^\mathcal{Y} \neq l_j^\mathcal{Y}$} \\
0 & \mbox{if $l_i^\mathcal{X} = l_j^\mathcal{X} \wedge l_i^\mathcal{Y} = 
l_j^\mathcal{Y}$.}\end{array}
\right.
\nonumber
\end{align}
We can rewrite Eq.~\ref{eq:energy_tree_seg} in a similar way, and they satisfy the submodular 
property according to the triangle inequality~\cite{KoZaTPAMI2004}. The label costs can be solved as in \cite{DeOsIsIJCV2012}.



\noindent \textbf{Parameters.} We manually explore the parameter space based on 
the pixel-level accuracy in a heuristic fashion. We first tune the parameters involved in the video-level labeling, then those involved in the segment-level labeling, and finally, those involved in GPM by running the bidirectional inference as in Sec.~\ref{sec:infer}.


\section{Experiments}
\label{sec:exp}

We evaluate our method on the recently released A2D dataset 
\cite{XuHsXiCVPR2015} and use their benchmark to evaluate the performance; 
this is the only dataset we are aware 
of that incorporates actors and actions together. We compare with the 
top-performing trilayer model benchmark, and
 two strong semantic image segmentation methods, AHRF 
\cite{LaRuKoTPAMI2014} and FCRF~\cite{KrKoNIPS2011}.
For AHRF, we use the public available code from \cite{LaRuKoTPAMI2014} as 
it contains a complete pipeline from training classifiers to learning and inference. 
For FCRF, we extend it to use the same features as our method.

\begin{table*}[t!]
\centering
\includegraphics[width=0.79\linewidth]{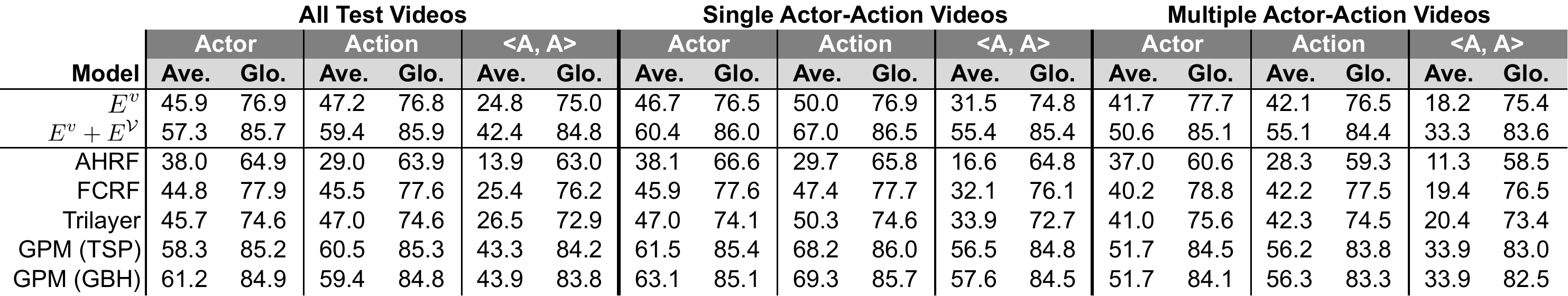}
\caption{The overall performance on the A2D dataset, where the performance is 
calculated for all test videos, single actor-action videos and multiple 
actor-action videos. The top two rows are intermediate results of full model 
(sub-parts of the energy). The middle three rows are comparison methods. The 
bottom two rows are our full model with different supervoxel hierarchies for 
the grouping process.}
\label{tab:exp_dataset}
\vspace{-2mm}
\end{table*}

\noindent \textbf{Data Processing.} We experiment with two distinct 
supervoxel trees: one is extracted from the 
hierarchical supervoxel segmentations generated by GBH \cite{GrKwHaCVPR2010}, 
where supervoxels across mutliple levels natively form a tree structure 
hierarchy, and the other one is extracted from mutliple runs of a generic 
non-hierarchical supervoxel segmentation by TSP \cite{ChWeIICVPR2013}. To 
extract a tree structure from the non-hierarchical video segmentations, we first 
sort the segmentations by the number of supervoxels they contain. Then we 
enforce the supervoxels in the finer level segmentation to have one and only one 
parent supervoxel in the coarser level segmentation, such that the two supervoxels 
have the maximal overlap in the video pixel space.  We use four levels from a 
GBH hierarchy, where the number of supervoxels 
varies from a few hundred to less than one hundred.
We also use four different runs of TSP to construct another 
segmentation tree where the final number of nodes contained in the tree varies 
from 500 to 1500 at the fine level, and from 50 to 150 at the coarse level.


We also use TSP to generate the segments for the base labeling CRF.
We extract the same set of appearance and motion features as in 
\cite{XuHsXiCVPR2015} (we use their code) and train one-versus-all linear SVM 
classifiers on the segments for three sets of labels: actor, action, and 
actor-action pair, separately. At the video-level, we extract improved dense 
trajectories \cite{WaScICCV2013}, and use Fisher vectors \cite{PeZoQiECCV2014} 
to train linear SVM classifiers at the video-level for the actor-action pair. We use the 
bidirectional inference and learning methods described in Sec.~\ref{sec:infer} 
and follow the train/test splits used in \cite{XuHsXiCVPR2015}. The output of 
our system is a full video pixel labeling. We evaluate the performance on 
sampled frames where the ground-truth is labeled. 


\begin{figure}[t!]
\centering
\includegraphics[width=0.99\linewidth]{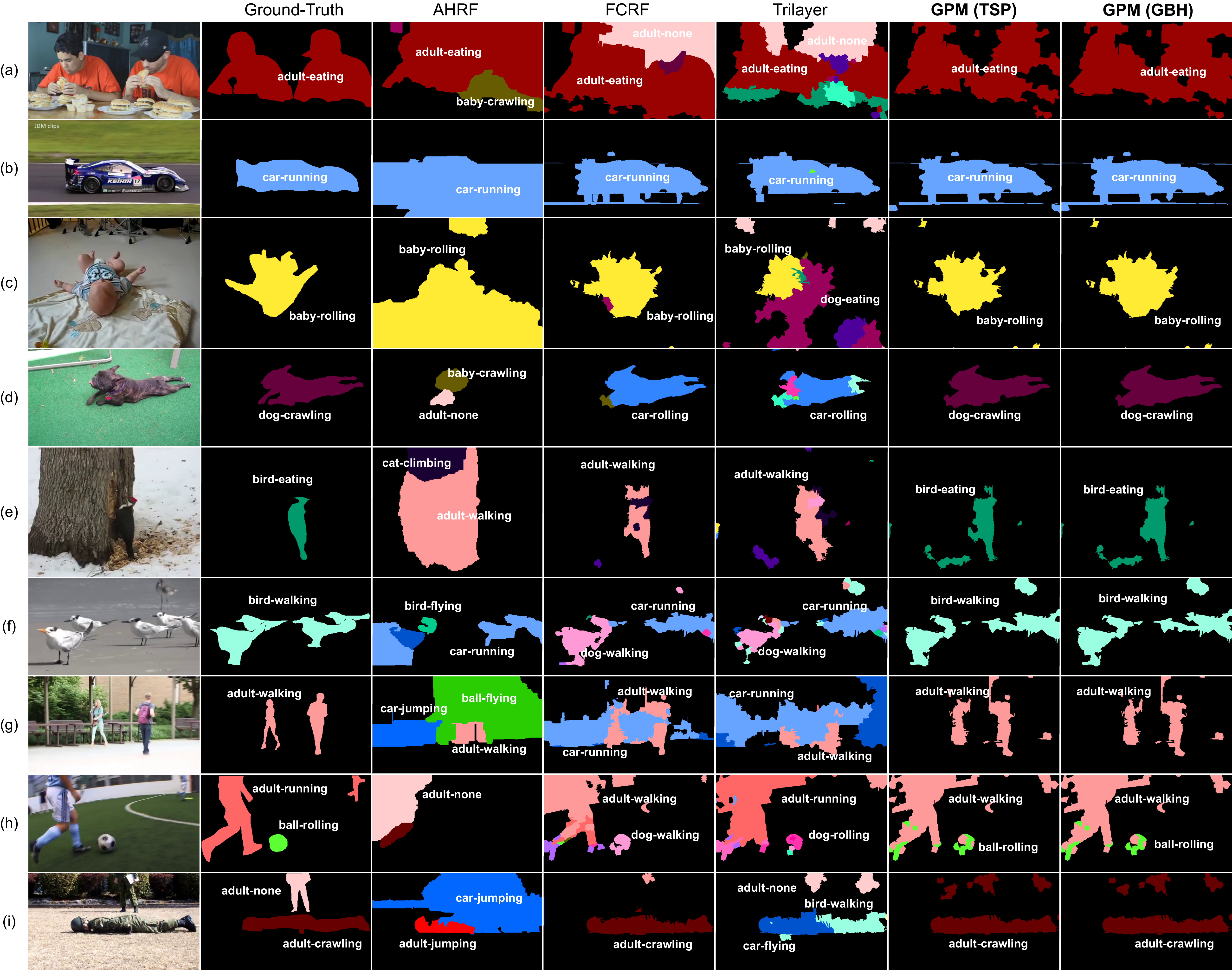}
\caption{Example video labelings of the actor-action semantic segmentations for all methods.
(a) - (c) are videos where most methods get correct labelings; (d) - (g) are 
videos where only our GPM models get the correct labelings; (h) - (g) are 
difficult videos in the dataset where the GPM models get partially correct 
labelings. Colors used are from the A2D benchmark \cite{XuHsXiCVPR2015}.}
\label{fig:visual_all}
\vspace{-2mm}
\end{figure}

\begin{table*}[t!]
\centering
\includegraphics[width=0.90\linewidth]{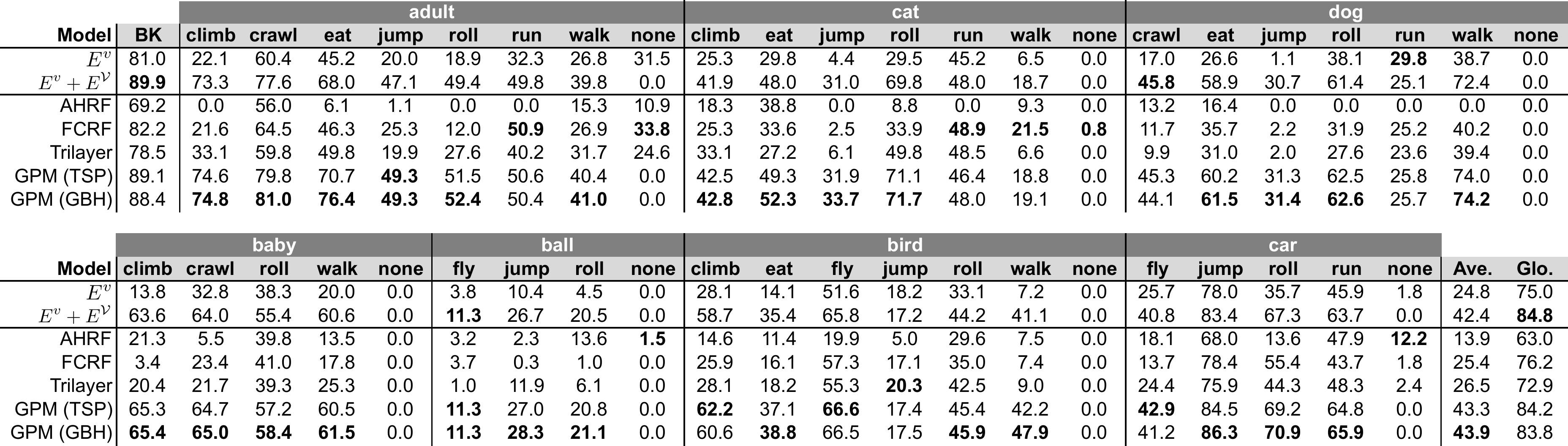}
\caption{The performance on individual actor-action labels using all test videos. The leading scores for each label are in bold font.}
\label{tab:exp_a2}
\vspace{-3mm}
\end{table*}

%

\noindent \textbf{Results and Comparisons.} We follow the 
benchmark evaluation in \cite{XuHsXiCVPR2015} and evaluate 
performance for joint actor-action and separate individual tasks.
Tab.~\ref{tab:exp_dataset} shows the overall results of all methods in three 
different calculations: when all test videos are used; when only videos containing 
single-label actor-action are used; and when only videos containing multiple 
actor-action labels are used. Roughly one-third of the videos in the A2D 
dataset have multiple actor-action labels. Overall, we observe that our 
methods (both GPM-TSP and GPM-GBH) outperform the next best one, the trilayer 
method, by a large margin of 17\% average per-class accuracy and more than 10\% 
global pixel accuracy over all test videos. The improvement of global pixel 
accuracy is consistent over the two sub-divisions of test videos, and the 
improvement of average per-class accuracy is larger on videos that only contain 
single-label actor-action. We suspect that videos containing multiple-label 
actor-action are more likely to confuse the video-level classifiers. 

We also observe that the added grouping process in GPM-TSP and GPM-GBH 
consistently improves the average per-class accuracy over the intermediate 
result ($E^v+E^\mathcal{V}$) on both single-label and multiple-label 
actor-action videos. There is a slight decrease on the global pixel accuracy. We 
suspect the decrease mainly comes from the background class, which contributes 
a large portion of the total pixels in evaluation. To verify that, we also 
show the individual actor-action class performance in Tab.~\ref{tab:exp_a2} when 
all test videos are used. We observe that GPM-GBH has the best performance on 
majority classes and improves $E^v+E^\mathcal{V}$ on all classes except 
\textit{dog-crawling}, which further shows the effectiveness of the grouping 
process. The performance of our method using the GBH hierarchy is slightly 
better than our method using the TSP hierarchy. 
We suspect that this is due to the GBH 
method's greedy merging process that complements
the Gaussian process in TSP, such that the resulting segmentation 
complements the segment-level TSP segmentation we used.

Figure \ref{fig:visual_all} shows the visual comparison of video labelings for 
all methods, where (a)-(c) show cases where methods output correct labels and 
(d)-(g) show cases where our proposed method outperforms other methods.
We also show failure cases 
in (h) and (i) where videos contain complex actors and actions. For example, our 
method correctly labels the \textit{ball-rolling} but confuses the label 
\textit{adult-running} as \textit{adult-walking} in (h); we correctly label 
\textit{adult-crawling} but miss the label \textit{adult-none} in (i).

\vspace{-2mm}
\section{Conclusion}
\label{sec:conclude}

Our thorough experiments on the A2D dataset show that when the segment-level 
labeling is combined with secondary processes, such as our grouping process 
models and video-level recognition signals, the semantic segmentation performance 
increases dramatically.
For example, GPM-GBH improves almost every class of actor-action labels compared 
to the intermediate result without the supervoxel hierarchy, i.e., without the 
dynamic grouping of CRF labeling variables. 
This finding strongly supports our motivating argument that the two sets of 
labels, actors and actions, are best modeled at different levels of 
granularities and that they have different space-time orientations in a video.

In summary, our paper makes the following contributions to the actor-action semantic segmentation problem:
\begin{enumerate}[labelsep=5pt, labelwidth=0pt,leftmargin=12pt,itemsep=0ex, 
parsep=0pt, topsep=0pt, partopsep=0pt]
\item A novel model that dynamically combines segment-level labeling 
with a hierarchical grouping process that influences connectivity of the 
labeling variables. 
\item An efficient bidirectional inference method that iteratively solves the 
two conditional tasks by graph cuts for labeling and binary linear programming 
for grouping allowing for continuous exchange of information.
\item A new framework that uses video-level recognition signals as cues for 
segment-level labeling thru multi-label labeling costs and the grouping process model.
\item Our proposed method significantly improves performance (60\% relative 
improvement over the next best method) on the recently released large-scale actor-action semantic video dataset \cite{XuHsXiCVPR2015}.
\end{enumerate}
Our implementations as well as the extended versions of AHRF and FCRF will be released upon publication.

\noindent \textbf{Future Work.} We set two directions for our future work. First, although our model is able to improve the segmentation performance dramatically, the opportunity of this joint modeling to improve video-level recognition is yet to be explored. Second, our grouping process does not incorporate semantics in the supervoxel hierarchy; we believe this would further improve results.

\clearpage
{\small
\bibliographystyle{ieee}
\bibliography{a2s2}
}

\end{document}